\documentclass{article}

\usepackage[preprint]{corl_2023} %

\usepackage{amsmath,amssymb,stmaryrd,mathtools}
\usepackage{bbm}
\usepackage{subfig} 
\usepackage{wrapfig}
\usepackage{xcolor}
\usepackage{xspace}

\definecolor{nice_green}{rgb}{0.01, 0.47, 0.08}
\definecolor{nice_blue}{rgb}{0, 0.41, 0.61}

\newcommand{\para}[1]{\smallskip \noindent \textbf{{#1}.}}

\newcommand{\example}[1]%
{
\textbf{\textit{Running example:}}
\textit{#1}
}

\newcommand{\change}[1]%
    {\textcolor{black}{#1}}

\newcommand{\strike}[1]%
    {}

\newcommand{\thor}{T}

\newcommand{\robot}{pursuer\xspace}

\newcommand{\agent}{evader\xspace}
\newcommand{\agents}{evaders\xspace}
\newcommand{\Robot}{Pursuer\xspace}
\newcommand{\Agent}{Evader\xspace}
\newcommand{\R}{\mathrm{p}}
\newcommand{\A}{\mathrm{e}}
\newcommand{\ang}{\theta}
\newcommand{\xR}{x^\R}
\newcommand{\xA}{x^\A}
\newcommand{\uR}{u^\R}
\newcommand{\uA}{u^\A}
\newcommand{\xrel}{x^{\mathrm{rel}}}
\newcommand{\pxrel}{p_x^{\mathrm{rel}}}
\newcommand{\pyrel}{p_y^{\mathrm{rel}}}
\newcommand{\threl}{\ang^{\mathrm{rel}}}
\newcommand{\dpxrel}{\dot{p}_x^{\mathrm{rel}}}
\newcommand{\dpyrel}{\dot{p}_y^{\mathrm{rel}}}
\newcommand{\dthrel}{\dot{\ang}^{\mathrm{rel}}}

\newcommand{\zhat}{\hat{z}}
\newcommand{\obs}{o}
\newcommand{\mean}{\hat{x}^\mathrm{rel}}
\newcommand{\cov}{\hat{\Sigma}}
\begin{document}
\title{Learning Vision-based Pursuit-Evasion Robot Policies}

\author{Andrea Bajcsy$^*$\And Antonio Loquercio\thanks{denotes equal contribution. All authors are affiliated to UC Berkeley.}%
\And Ashish Kumar\And Jitendra Malik}%

\makeatletter
\let\@oldmaketitle\@maketitle%
\renewcommand{\@maketitle}{\@oldmaketitle%
\centering
\vspace{-0.2in}
\includegraphics[width=\textwidth]{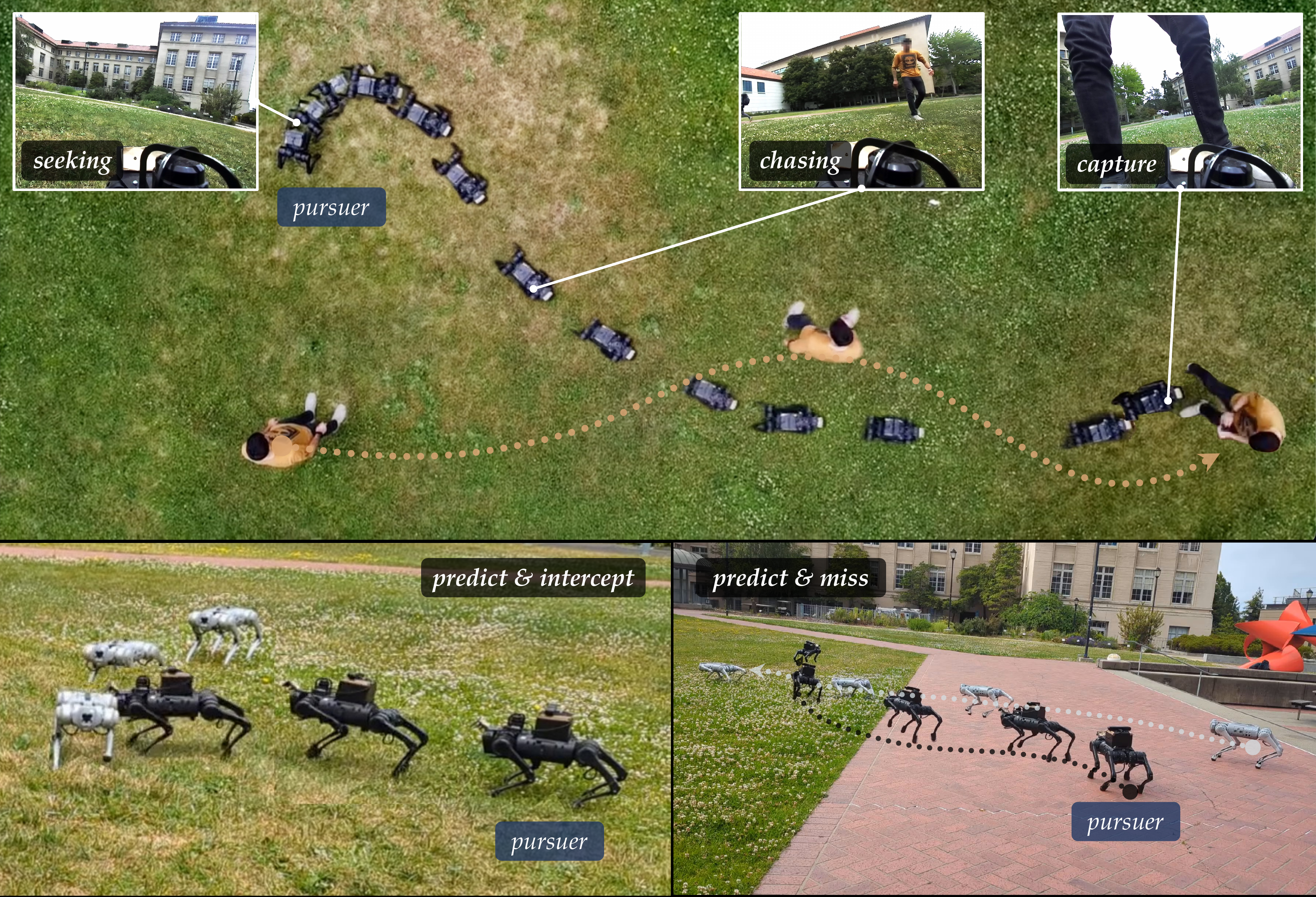}
\captionof{figure}{Our approach deployed in a pursuit-evasion interaction in the wild. Our policy (black robot acting as pursuer) automatically synthesizes behaviors like slowing down and information gathering, accelerating upon detection, and prediction and interception. Video results at \href{https://abajcsy.github.io/vision-based-pursuit/}{\texttt{https://abajcsy.github.io/vision-based-pursuit/}}.}
\label{fig:front figure}
\vspace{-0.1in}
\bigskip}
\makeatother
\maketitle

\begin{abstract}
    Learning strategic robot behavior---like that required in pursuit-evasion interactions---under real-world constraints is extremely challenging.
    It requires exploiting the dynamics of the interaction, and planning through both physical state and latent intent uncertainty. In this paper, we transform this intractable problem into a supervised learning problem, where a fully-observable robot policy generates supervision for a partially-observable one. 
    We find that the quality of the supervision signal for the partially-observable \robot policy depends on two key factors: the balance of diversity and optimality of the \change{\agent's} behavior, and the strength of the modeling assumptions in the fully-observable policy. 
    We deploy our policy on a physical quadruped robot with an RGB-D camera on pursuit-evasion interactions in the wild.
    Despite all the challenges, the sensing constraints bring about creativity: the robot is pushed to gather information when uncertain, predict intent from noisy measurements, and anticipate in order to intercept.
\end{abstract}


\section{Introduction}
\label{sec:intro}

Robot learning has accelerated progress for embodied agents acting ``in the wild'':
quadrupedal and wheeled robots navigate through hard-to-model terrains \cite{lee2020learning, frey2023fast, kumar2021rma, loquercio2022learn, pan2017agile}, quadrotors fly at their limits \cite{loquercio2021learning}, and robotic arms deftly manipulate deformable objects \cite{chi2022iterative}.
However, these successes are limited to robots acting in isolation; in reality, robots deployed at scale will inevitably interact with other agents, like people or other robots.
    
In-the-wild multi-agent interactions raise significant challenges: not only does a robot have to account for perception-induced uncertainty of the physical state (e.g., ego state, positions of others), but it must also account for uncertainty in other agents’ future behavior.
This problem setting is traditionally modeled by decentralized partially-observable Markov decision processes (Dec-POMDPs) or partially-observable stochastic games (POSGs). While in theory, solutions to these formulations would automatically yield desirable behaviors like information gathering when uncertain, in practice they are notoriously intractable.   %

Nevertheless, human and animal behavior exhibits these abilities \cite{wilson2013locomotion}. 
Pursuit-evasion interactions are a canonical example: the \robot gathers information about the hidden \agent by turning and scanning the environment; upon detection, the \robot has to continuously strategize about its next move without perfect knowledge of how the \agent will react, all from onboard sensors.  
In this work, we take the first steps towards building similar capabilities into autonomous robots.

Our key idea is to leverage a fully-observable policy to generate supervision for a partially-observable one. 
However, the classic paradigm of privileged learning \cite{chen2020learning} does not apply naively to this setting.
Namely, privileged information depends not only on the robot, but also on the other agent's behavior, which is dictated by \textit{more} than just physics; it is dictated by the other agent's intent. 
Therefore, we design a learning procedure to first build a low-dimensional latent representation of intent from future \agent trajectories and then learn to estimate this representation
from a history of \robot actions and observations.

Through extensive empirical analysis, we find that the quality of the supervision signal depends on a delicate balance between the diversity of the agents' behavior and the optimality of the interaction. 
In addition, there are many models for generating the fully-observable supervisor policy (e.g., game-theory \cite{bacsar1998dynamic}, multi-agent RL \cite{gronauer2022multi}), each with their own potential strengths and weaknesses. 
\change{We discover that fully-observable policies obtained under strong modeling assumptions (e.g., both agents play under perfect-state Nash equilibrium), are less effective at supervising partially observable ones.}

Informed by this analysis, we synthesize a policy that \textit{automatically} takes actions to resolve physical state uncertainty (e.g., looking around to see detect where the other agent is) while also generating predictions about other agents' intent to yield strategic behavior. We deploy this policy on a physical legged robot in a pursuit-evasion game, where it interacts with humans or other legged robots (Fig.~\ref{fig:front figure}). Note that the robot only uses onboard sensing, e.g., proprioception and an RGB-D camera, to estimate its state and other agents' physical state and intent.

\section{Related Work}
\label{sec:related_work}

\noindent\textbf{Dynamic Games \& Multi-Agent RL.} 
Dynamic game theory has a long history of modeling strategic interaction between multiple agents \cite{isaacs1954differential, isaacs1999differential, shapley1953stochastic, littman1994markov} and has influenced fields like robust control \cite{mitchell2005time} 
and reinforcement learning \cite{zhang2021multi, pinto2017robust, bucsoniu2010multi}. 
Both traditional and modern variants of dynamic games have predominantly assumed knowledge of perfect state. While this has been successful in contexts like robustness to physical disturbances \cite{pinto2017robust, tang2020learning, hsu2022isaacs}, it's a limiting assumption for real-world interaction. 
Partially observable stochastic games provide a mathematical model of strategic interaction under partial observability \cite{zhang2021safe},
where all players have only partial information about environment state. 
However, they are tremendously computationally expensive to solve, and approximations are an active area of research~\cite{schwarting2021stochastic}.
To make the optimization tractable, multi-agent reinforcement learning (MARL) algorithms exploit large-scale simulation and neural network representations~\cite{lowe2017multi, pinto2017robust, woodward2020learning, foerster2018counterfactual, kim2022influencing}. Such approaches have achieved impressive results in simulation interactions like hide-and-seek \cite{baker2019emergent}, video games like Starcraft \cite{vinyals2019grandmaster}, diplomacy \cite{meta2022human}, and board games like Go \cite{silver2016mastering}. 
However, to-date, multi-agent RL approaches have not yet scaled to embodied systems acting under real-world sensing constraints.
Although these approaches cannot be applied to our physical system out-of-the-box, we do take inspiration from this line of work in the design of privileged fully-observable \robot policy and a highly strategic \agent policy.

\para{Latent Intent Modeling} 
Recent works \cite{xie2021learning, parekh2022rili, wang2022influencing} learn a latent representation of agent intent via reconstructing a dataset of fully-observed states and rewards. This line of works assumes that the latent intent changes only \textit{between} interaction episodes and not during an interaction.
To handle intent changes \textit{during} interaction \cite{he2023learning} learns an estimator of a human's latent state to predict their immediate next action. These works overwhelmingly assume that the only hidden state in the interaction is the opponent's intent: the physical state of the robot, the opponent's state, and possibly the opponent's action are assumed to be observable. We address the constraints imposed by on-board robot perception, where the \agent's physical state, latent intent, and action are all hidden.

\para{Multi-Quadruped Interaction}
Progress in low-level control for quadrupedal robots \cite{lee2020learning, kumar2021rma,loquercio2022learn} has increased the interest in combining low-level controllers with high-level decision-making \cite{huang2022creating}. 
However, multi-agent quadruped interactions have been relatively under-explored.  
Most relevant is \cite{nachum2019multi} which trains a \textit{centralized}  high-level coordination policy  with perfect (global) state for two robots pushing a box. 
\cite{fawcett2022distributed, kim2023safety} present cooperative control of robots via model predictive control and control barrier functions. 
To the best of our knowledge, our work is the first to demonstrate autonomous interaction between a quadruped and another robotic or human agent truly in the wild.

\section{Overview}
We seek a robot policy that can strategically interact with another agent in a decentralized fashion (i.e., no explicit communication) and only using proprioception and a single onboard RGB-D camera.
Although our technical approach is general, we ground this work in pursuit-evasion games~\cite{isaacs1954differential}, which \change{exhibit core challenges at the heart of real-world multi-agent interaction: partial observability, nonlinear physical dynamics (e.g., quadrupedal dynamics), low-latency decision-making, and a need for strategic planning.} 

\begin{figure}[t!]
    \centering
    \includegraphics[width=0.85\textwidth]{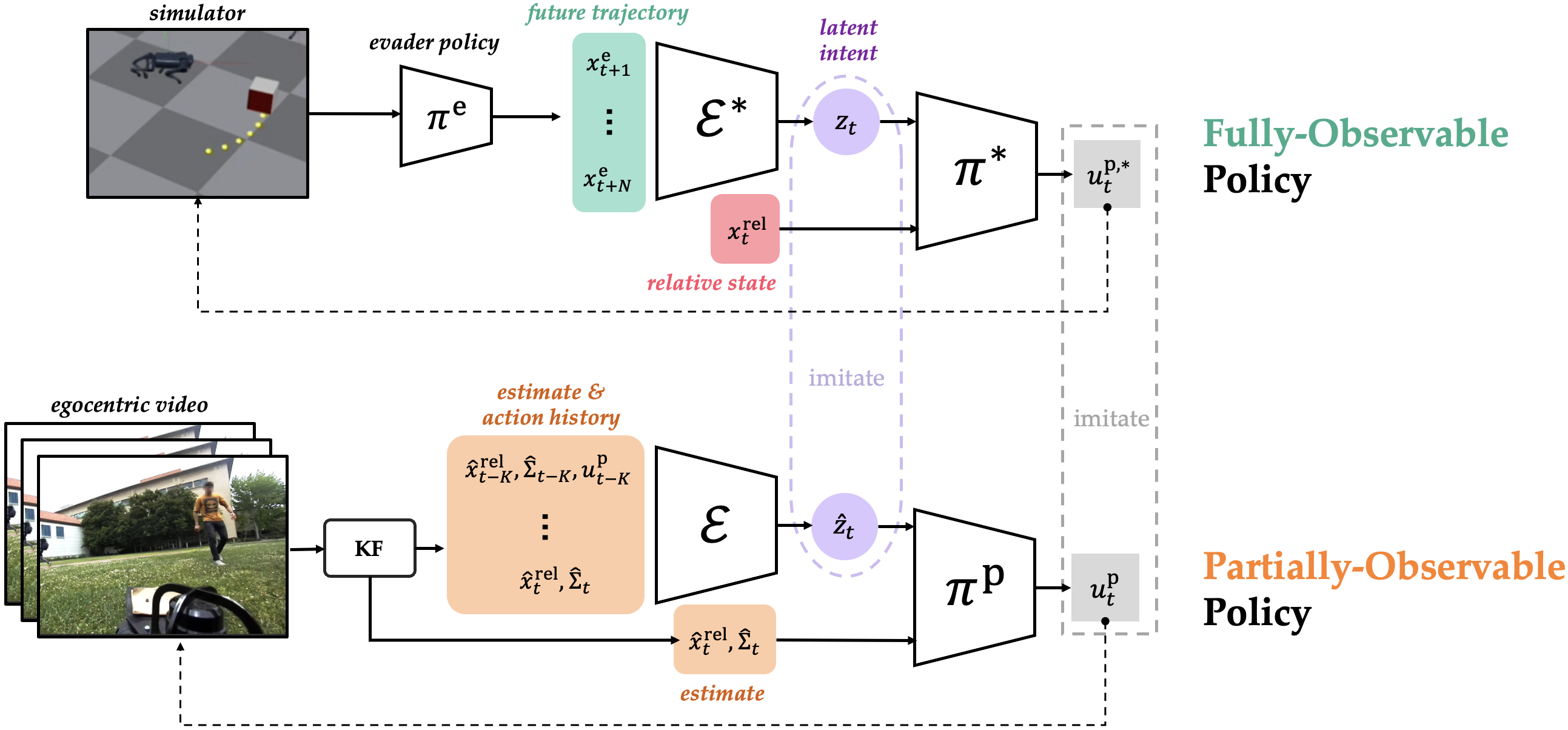}
    \caption{(top) The fully-observable policy knows the true relative state and gets privileged access to the future \agent trajectory from which it learns a representation the \agent intent. (bottom) The partially-observable policy must plan through physical and latent intent uncertainty.}
    \label{fig:approach}
    \vspace{-1em}
\end{figure}

We approach this problem using privileged learning \cite{chen2020learning}. We found that directly using reinforcement learning to train a policy that reasons \textit{strategically} through \textit{partial observability} was unsuccessful. 
The key to our approach is to leverage a fully-observable policy to generate supervision for the partially-observable one. During privileged training, we leverage a new type of privileged information: the future state trajectory of the \agent. 
We first learn a \textbf{fully-observable policy ($\pi^*$)} (top, Fig.~\ref{fig:approach}), that gets access the true future $N$-step state trajectory of the \agent and the current true relative state.
This enables $\pi^*$ to quickly learn actions that account for the \agent's behavior 
by using a learned latent intent, $z_t$, that encodes the future trajectory of the other agent.

We then distill this policy into a \textbf{partially-observable policy ($\pi^\R$)} which only uses an egocentric video stream from an onboard RGB-D camera (bottom, Fig.~\ref{fig:approach}). Specifically, $\pi^\R$ gets access to a \textit{history} of relative state \textit{estimates and uncertainties} which are generated via a standard Kalman Filter \cite{kalman1960new}. 
Even though the Kalman filter is an incredibly coarse approximation of the true system, the uncertainty information captured by the covariance matricies is sufficient for the prediction policy to learn information-gathering behaviors (like turning and looking for the \agent), when combined with the teacher policy. 
In this light, our privileged learning approach can be viewed as an approximation to the optimal policy obtained via solving the underlying, but intractable, decentralized partially-observable Markov decision process (Dec-POMDP). 

Our partially-observable policy can be applied zero-shot in the real world using the output of an off-the-shelf object detector \cite{zed2023stereolabs}. 
We deploy it in the wild to play a pursuit-evasion game with a human \agent and another quadrupedal robot controlled by a human operator.

\section{Approach}
\label{sec:approach}

Given an \agent policy $\pi^\A$, we define the 
\robot's planning problem as a finite-horizon, discrete-time optimization problem.  
We seek a policy for the \robot $\pi^\R: \mathcal{O}^\R \rightarrow \mathcal{U}^\R$ which maps from observations to actions that maximizes the following objective:
\begin{equation}
\label{eq:planning_problem}
    J(\pi^\R, \pi^\A) = \mathbb{E}_{\tau \sim p(\tau \mid \pi^\R, \pi^\A)} \Big[ \sum^{\thor}_{t=0} \gamma^t r_t \Big].
\end{equation}
Here, $\tau = \{(\xR_0, \xA_0, \uR_0, \uA_0, \obs^\R_0, \obs^\A_0, r_0) \hdots (\xR_T, \xA_T, \uR_T, \uA_T, \obs^\R_T, \obs^\A_T, r_T)\}$ is the joint trajectory of states, actions, observations, and rewards induced by a pair of \robot and \agent policies, drawn from the distribution $p(\tau \mid \pi^\R, \pi^\A)$. The discount factor is denoted by $\gamma$. More formally, this optimization defines the solution to a two-agent, finite horizon decentralized partially-observable Markov decision process (Dec-POMDP). 

We denote the global physical state of the \robot as $\xR \in \mathbb{R}^{n_\R}$ and the \agent to be $\xA \in \mathbb{R}^{n_\A}$. Note that the \robot policy $\pi^\R$ does not observe the global state of the agents. 
The robot's high-level linear and angular velocity commands are denoted by $\uR \in \mathcal{U}^\R$ and the \robot's low-level joint torques are controlled via a pre-computed walking policy. 
The \agent also controls its linear and angular velocity, $\uA \in \mathcal{U}^\A$. Both $\mathcal{U}^i, i \in \{\R, \A\}$ are bounded sets, modeling actuation limits. \change{For example, in simulation, the maximum linear speed of the \robot is 3 m/s, and the \agent is 2.5 m/s.}

The \robot is rewarded for minimizing the distance between the two agents at each timestep, and obtains a termination bonus upon capture:
\begin{equation}
    \textit{Pursuit: } r_t = -||\xA_t - \xR_t||_2^2, \quad \textit{Capture: } r_T = \alpha \cdot \mathbbm{1}\{||\xA_T - \xR_T||_2^2 \leq 0.8\}
\end{equation}

\change{\para{Asymmetries} Our setting has three asymmetries that induce complexity: 1) \textit{information} (agents have limited FOV and partial state), 2) \textit{dynamics} (e.g., robot quadruped interacting with human), and 3) \textit{control bound} asymmetry (e.g., agents with different maximum speeds).}

\para{Ego-Centric State} 
All agents reason about the \textit{relative} physical state in their own body frame. 
In a slight abuse of notation, we refer to $\xrel := [p_x^\mathrm{rel}, p_y^\mathrm{rel}, \theta^{\mathrm{rel}}]^\top$ as the true relative planar position and orientation of the exo-centric agent in the ego-centric agent's body frame.

\para{State Estimation} In the wild, the true relative state is not available due to sensing limits. Instead, the \robot estimates $\xrel_t$ from the output of a  3D object detector using the RGB camera \cite{zed2023stereolabs}.
Let $\obs_t^\R \in \mathcal{O}^\R$ be the 3D relative position of the \agent with respect to the \robot's camera frame\footnote{If the \agent is out of the field of view, then $\obs_t^\R = \emptyset$ and no measurement update is performed.}. The \robot's relative state estimate are the mean and covariance of a Kalman filter: $(\mean_t, \cov_t) = \mathrm{KF}(\obs^\R_{0:t}, \uR_{0:t})$. While a suite of more complex filter designs could be used for even better performance \cite{lee2020multimodal}, we find that using an unoptimized Kalman filter is sufficient for the robot policy to learn information-gathering behaviors. 

\para{\Agent Policy} 
The \agent policy is key for enabling the \robot to learn strategic maneuvers. However, where does the \agent policy come from? Datasets of quadrupeds interacting with other agents in the wild do not exist, and simulating human-robot or robot-robot interactions that capture the diversity of the real-world is an ongoing challenge for simulation-based robotics.
Instead, we take an investigative approach and study three simulated \agent policy models: random motion primitives, multi-agent RL, and dynamic game theory. Across all models, we assume the \agent has access to the current true relative state in their own body frame. Details are in Sec.~\ref{subsec:ablation_opponent_policy}.

\subsection{Fully-Observable Policy: Teacher}

To learn the \robot teacher policy $\pi^*$, we must address the challenge that privileged information depends on the \agent's behavior which is dictated by their dynamics and intent.

\textbf{Future \Agent Trajectory \& Latent State.} The fully-observable policy $\pi^*$ gets access to both the 
true \robot relative state, $\xrel_t$, and the future $N$ states of the evader: $x^\A_{t:t+N}$. 
Since the \robot reasons in a relative coordinate system, the \agent trajectory is converted into the \robot's body frame starting from state at the start of the prediction horizon. 
This relative state trajectory, $\xrel_{t:t+N} \in \mathbb{R}^{N \times 3}$, is input into an encoder, $\mathcal{E}^*(\xrel_{t:t+N}) = z_t \in \mathbb{R}^8$ which learns
a low-dimensional latent representation.
Intuitively, $z_t$ should capture low-dimensional information about the \agent's near-term behavior: for example, the \agent's goal direction, their policy class (e.g., spline coefficients), or control bounds. At each timestep, $z_t$ is re-inferred. 

\textbf{Design.} Although this \robot policy is clearly not deployable in the real world, it enables us to convert the intractable planning problem in Eq.~\ref{eq:planning_problem} to a Markov Decision Process (MDP), amenable to off-the-shelf RL methods \cite{schulman2017proximal}. We use Proximal Policy Optimization \cite{schulman2017proximal} for training. 
The policy $\pi^*$ and the privileged encoder $\mathcal{E}^*$ are both three-layer MLPs with $[512, 256, 128]$ hidden units.

\subsection{Partially-Observable Policy: Student}
The partially-observable policy, $\pi^\R$, relies on RGB camera observations $o^\R_t \in \mathcal{O}^\R$. We use a off-the-shelf 3d object detector 
\cite{zed2023stereolabs} to convert from the raw RGB image observable $o^\R_t$ to an detected relative position and heading in the \robot's camera frame, $y_t \in \mathbb{R}^3$. We use a Kalman Filter to generate an estimated relative state $\mean_t$ and an associated covariance $\cov_t$. 
We use a simplified state transition model $\mean_{t+1} = A\mean_t + B\uR_t$ which ignores the role of the \agent (i.e., $\uA_t \equiv 0$) during the the prediction\footnote{\change{This removes the need for a velocity estimator which can be hard to design and noisy in reality. While this makes filtering imperfect, we empirically find that the learned policy can compensate for inaccuracies.}} step \change{(details in Appendix~\ref{app:kalman_filter})}.
The history of relative state estimates and \robot actions are encoded into the estimated lower-dimensional latent intent $\mathcal{E}(\mean_{0:t}, \cov_{0:t}, u^\R_{0:t-1}) = \zhat_t$.

\para{Design} We use DAGGER \cite{ross2011reduction} and the fully-observable policy $\pi^*$ to supervise both the latent intent estimate and the action at each timestep. The policy network is a 3-layer MLP with $[512, 256, 128]$ hidden units, and the encoder $\mathcal{E}$ is a 1-layer LSTM with hidden state $256$.

\begin{figure}[t!]
    \centering
    \includegraphics[width=\textwidth]{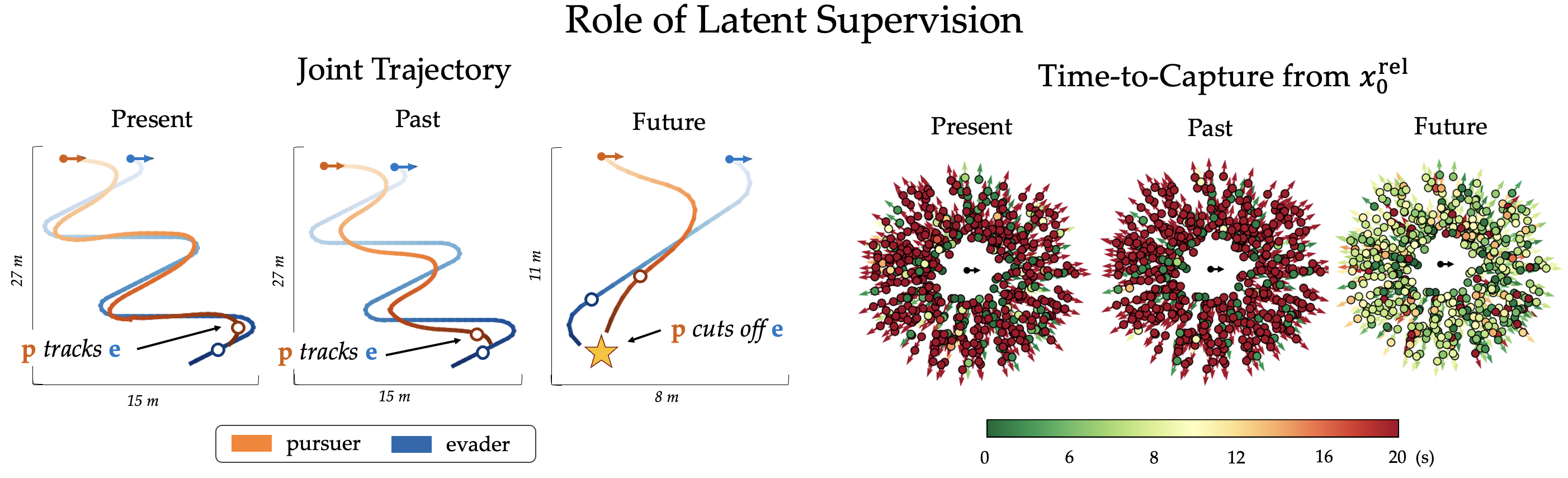}
    \caption{(left) Joint trajectories reveal that using only the present state or directly a history of the past states leads to sub-optimal performance. Supervision from the future enables prediction and fast capture. (right) Time-to-capture as a function $\xrel_0$: future supervision quarters the capture time. }
    \label{fig:react_v_post_v_pred}
    \vspace{-1em}
\end{figure}

\section{Simulation Experiments}
\label{sec:simulation}

\change{We first want to understand the design choices that are important to learn a successful \robot policy: 
1) the ability to learn strategic behavior, 
2) the \agent policy $\pi^\A$ that the robot interacts with at \textit{training} time and 3) the \agent interaction at \textit{deployment} time. 
We perform a set of simulation experiments to ablate the design of the \robot policy (Sec.~\ref{subsec:ablation_latent_state}), study the effect of the \agent on distillation (Sec.~\ref{subsec:ablation_opponent_policy}), and analyze test-time adaptation of the \robot policy to out-of-distribution opponents (Sec.~\ref{subsec:adapting_to_ood}). }
We use Isaac Gym~\cite{rudin2022learning} for training and evaluation, and report results over 500 random initial conditions.

\subsection{Predictive representations enable strategic behavior}
\label{subsec:ablation_latent_state}
\change{One of the key types of privileged information we leverage is future trajectory state, which leaks information about the future intent of the other agent. In this section, we ask the question \textit{``What is the value of learning predictive representations for action?''} }
\change{Here, }we fix the \agent policy to be highly predictable and investigate alternative approaches to inferring the \agent's latent intent. The \agent always moves in Dubins' paths \cite{dubins1957curves} with a fixed time duration for turning or going straight determined randomly upon the start of the episode (\change{details in Sec.~\ref{sec:dubins} of the appendix}).
If the \robot has a high-quality understanding of the \agent's latent intent, it should be able to intercept it along its weaving path. Throughout this section, all policies observe \textit{ground-truth} relative states but not the \agent's latent intent.

We consider three approaches: a \textbf{reactive} \robot policy which only observes the present relative state and does not infer any latent intent, a \textbf{lookback} policy which must estimate the  latent intent from a history of relative states \textit{without supervision from the future}~\cite{he2023learning, xie2021learning} and a \textbf{lookahead} policy \change{(ours)}, which uses a history of relative states \change{and supervision from the future to} predict a latent representation of the \agent's future trajectory.
The \change{\textbf{lookback} and \textbf{lookahead} policies} use identical LSTM architectures for intent estimation.

The \textbf{reactive} policy fails to predict the \agent's behavior and is unable to do better than tracking the \agent and trailing behind it (left, Fig.~\ref{fig:react_v_post_v_pred}).
While adding a history improves the \robot's strategy (center, Fig.~\ref{fig:react_v_post_v_pred}), it still struggles to estimate the \agent's intent reliably.
In contrast, the \textbf{lookahead} policy, trained to predict a latent representation of the \agent's future trajectory, learns effective predictive behaviors (right, Fig.~\ref{fig:react_v_post_v_pred}).
\change{In addition, the \textbf{lookahead} policy converges 10 times faster than the \textbf{lookback} one (see Appendix Fig.~\ref{fig:lb_vs_la_train}). Overall, our experiments show that using the future trajectory as privileged information favors training and enables strategic behaviors.}

\begin{figure}[t]
    \centering
    \includegraphics[width=0.9\textwidth]{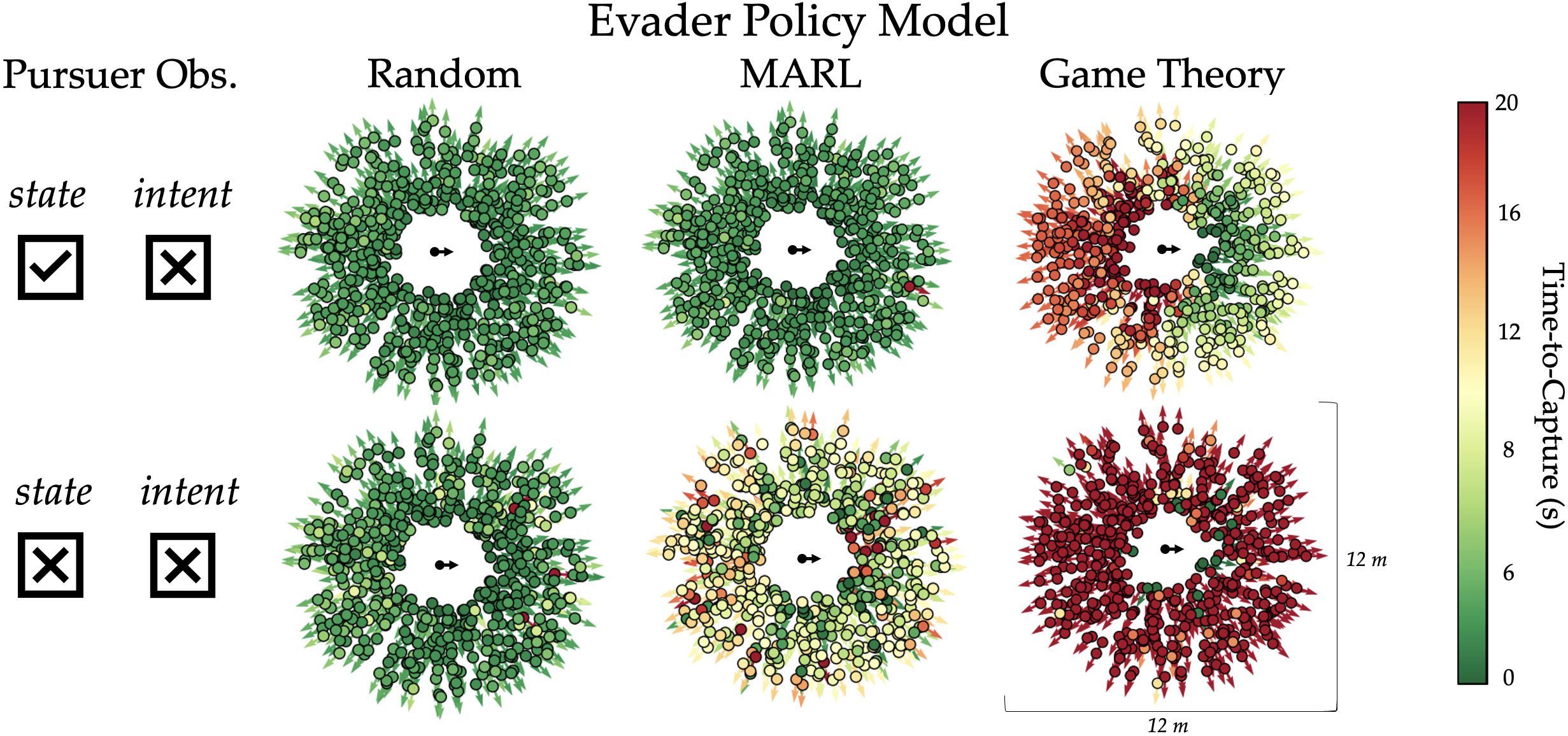}
    \caption{500 randomly sampled $\xrel_0$ initial positions and relative orientation. Colors indicate the normalized time-to-capture starting from the shown initial condition. (top row) Policy knows perfect physical state but infers latent intent from history of relative states and \robot actions, trained with three different \agent models: heuristic, MARL, and zero-sum game-theoretic. (bottom row) Partially-observable state and intent policy is supervised by the corresponding policy above.}
    \label{fig:full_vs_partial_obs}
    \vspace{-1em}
\end{figure}

\subsection{Distillation depends on balance of interaction diversity and optimality}
\label{subsec:ablation_opponent_policy}

\change{\para{Influence of \agent model on \robot policy} }
\change{Now that we have a teacher policy architecture, we turn to the role of the \agent on the teacher \robot policy.
We compute three fully-observable teacher policies, $\pi^*$, trained against three different \agent models, $\pi^\A$.}
With a slight abuse of notation, let $\xrel$ be the relative state in the \agent's body frame. 
The \textbf{random} 
\agent, $\pi^\A_{\mathrm{rand}}(t)$, randomly samples a set of controls to apply each 1-3 seconds. The \textbf{multi-agent RL} \agent, $\pi^\A_{\mathrm{marl}}(\xrel_t)$, is trained to evade a pre-trained \robot policy, equivalent to a single iteration of \cite{pinto2017robust}. 
Finally, \change{assuming perfect} relative state, our setting could be modelled by a zero-sum \textbf{game theory} model, whose solution characterizes the optimal pair of policies for the \robot and the \agent \cite{mitchell2005time}. We compute  $\pi^\A_{\mathrm{game}}(\xrel_t)$ via an off-the-shelf dynamic game solver \cite{mitchell2008flexible}. \change{Details on all \agents in Sec.~\ref{app:evader_design}. Top row in Fig.~\ref{fig:full_vs_partial_obs} shows that with perfect state, the \robot capture time is indistinguishable between \textbf{random} and \textbf{MARL} \agents, while the optimal \textbf{game theory} \agent maximally exploits the interaction. } 

\begin{wrapfigure}{r}{0.45\textwidth}
  \begin{center}
\includegraphics[width=0.4\textwidth]{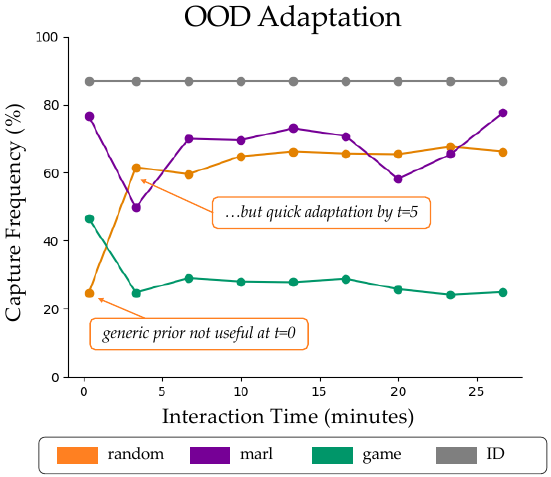}
  \end{center}
  \caption{\change{Coverage is more helpful than specialization for quick adaptation.}}
  \label{fig:continual_learn}
  \vspace{-2em}
\end{wrapfigure}

\change{\para{Distillation to partially-observable policy}} After training the fully-observable \robot policy against each \agent, we supervise the corresponding partially-observable \robot policy (bottom row, Fig.~\ref{fig:full_vs_partial_obs}). 
We find that the \textbf{game-theoretic} \robot policy makes for a poor supervisor because the supervision and interaction data operate under a perfect state-feedback Nash equilibrium assumption that is too hard to satisfy for the partially-observable policy.
\change{In contrast, robots trained against noisily-optimal (\textbf{MARL}) or extremely diverse (\textbf{random}) \agents have smaller in-distribution performance drops. This indicates that the interaction assumptions under which the teacher policy is obtained must be feasible for the partially-observable student. }

\subsection{For adaptation, coverage is more helpful than specialization}
\label{subsec:adapting_to_ood}

\change{Ultimately, the test-time distribution of the \agent is unknown a priori. Thus, we ask \textit{``How quickly and how effectively can different partially-observable \robot policies adapt to an out of distribution \agent?''} }
We simulate interaction between $\pi^\R_{\mathrm{rand}}, \pi^\R_{\mathrm{marl}}, \pi^\R_{\mathrm{game}}$ and the highly predictable Dubins' agent (Sec.~\ref{sec:dubins}). None of the pursuers were trained on this behavior. 
We collect batches of joint state trajectories, and then finetune the weights of the \robot's encoder $\mathcal{E}$ by supervising the latent $\zhat_t$ at each timestep via the privileged encoder $\mathcal{E}^*$. 
Since $\pi^\R_{\mathrm{rand}}$ has a generic prior on the \agent motion, the representation in $\mathcal{E}$ is rich enough to triple its capture frequency in just 5 min. (Fig.~\ref{fig:continual_learn}). \change{$\pi^\R_{\mathrm{marl}}$ is good at the start, but, due to its prior on the evader motion, it is less flexible and needs more data to see improvements. $\pi^\R_{\mathrm{game}}$, with a stronger prior than $\pi^\R_{\mathrm{marl}}$, fails to adapt and reaches a sub-optimal performance with the limited data. 
This indicates that to quickly adapt to agents ``in the wild'', which are neither random nor optimal, coverage is more helpful than specialization.}

\section{Real-World Results}
\label{sec:hardware}

Real-world interactions are out-of-distribution for two main reasons: (1) the behavior of the \agent is unscripted and possibly very different to what was observed in simulation, and (2) the physical dynamics of the \agent do not follow the unicycle model as in simulation.
We run two sets of experiments to study how our policies react to such conditions.

\begin{figure}[t!]
    \centering
    \includegraphics[width=0.98\textwidth]{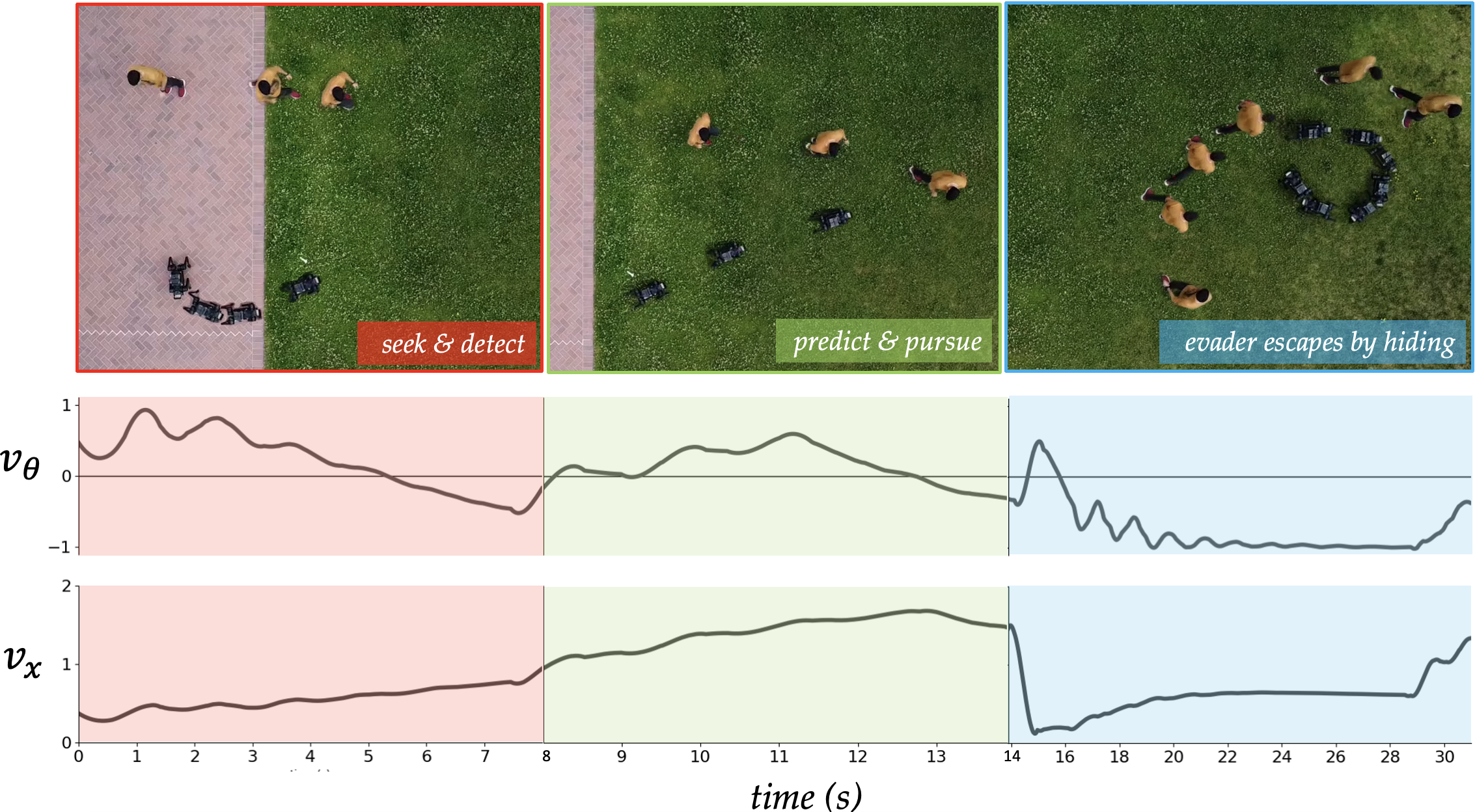}
    \caption{Single-shot interaction between a vision-based \robot policy and a human. \textcolor{red}{\textbf{Left:}} Since the human starts outside the FOV of the robot, the latter turns and seeks until it gets the first detection. \textcolor{nice_green}{\textbf{Center:}} The robot predicts the human will go straight and gallops to where the person will be. \textcolor{nice_blue}{\textbf{Right:}} Human strategically hides outside robot's FOV to escape.}
    \vspace{-1em}
    \label{fig:hardware_experiments}
\end{figure}

First, we ablate the pursuer policy and deploy \change{$\pi^\R_{\mathrm{rand}}$, $\pi^\R_{\mathrm{marl}}$, $\pi^\R_{\mathrm{game}}$} on a physical quadruped robot to interact with a human.
We observe that $\pi^\R_{\mathrm{rand}}$ and \change{$\pi^\R_{\mathrm{marl}}$} perform qualitatively similarly.
They showcase information-seeking motions when the \agent is not in the field of view and predictive strategies when the \agent is visible, i.e., heading towards where the \agent \emph{will be}, not where \emph{it is} (Fig.~\ref{fig:hardware_experiments}).
However, \change{$\pi^\R_{\mathrm{marl}}$} shows slightly better anticipation and faster reaction times.
\change{Conversely, $\pi^\R_{\mathrm{game}}$ shows inefficient information-seeking motions, taking long detours to reach the \agent.
Such performance, inline with the experiments from Sec.~\ref{sec:simulation}, confirms that the diversity of interaction data collected by a game-theoretic supervisor is not high enough to be robust to real-world interactions. }

Second, we keep the pursuer policy fixed and ablate the dynamics of the evader.
Concretely, we compare the performance of $\pi^\R_{\mathrm{rand}}$ when interacting against a human or another quadruped teleoperated by a human (Fig.~\ref{fig:front figure}).
In both cases, we observe aspects of strategic behavior. However, such behavior is more apparent during interaction with another robot. This is due to the robot's dynamics being closer to the unicycle model the policy was trained on in simulation. See website for videos.

\section{Discussion}

\para{Limitations} \change{Our current approach does not model the affordances of the environment, like obstacles, that could be strategically used by the \robot. }
However, to achieve this, the \robot needs to sense the environment geometry, potentially making optimization much more challenging.
\change{Additionally, the limited FOV assumption could be alleviated with different sensor designs (e.g., high-resolution 360 FOV camera), but introduce additional challenges like the computational burden of processing high-resolution images, that are interesting future work.}

\para{Conclusions}
Our paper takes the first steps toward learning vision-based robot policies that can reason strategically through partially-observable physical state and latent intent. 
We find interesting \robot behaviors when deployed on a physical quadruped robot with an RGB-D camera: information gathering under uncertainty, intent prediction from noisy state estimates, and anticipation of agents' motion.

\clearpage

\acknowledgments{This work was supported by the DARPA Machine Common Sense program and by the ONR MURI award N00014-21- 1-2801. The authors thank Noemi Aepli for her help with real-world experiments.}

\bibliography{references}  %

\clearpage 

\appendix
\appendix
\section{Appendix}

\change{\subsection{\Robot capture time as a function of train vs. test \agent}
\label{app:strategic_to_rand}}

\begin{figure}[h!]
    \centering
    \includegraphics[width=0.8\textwidth]{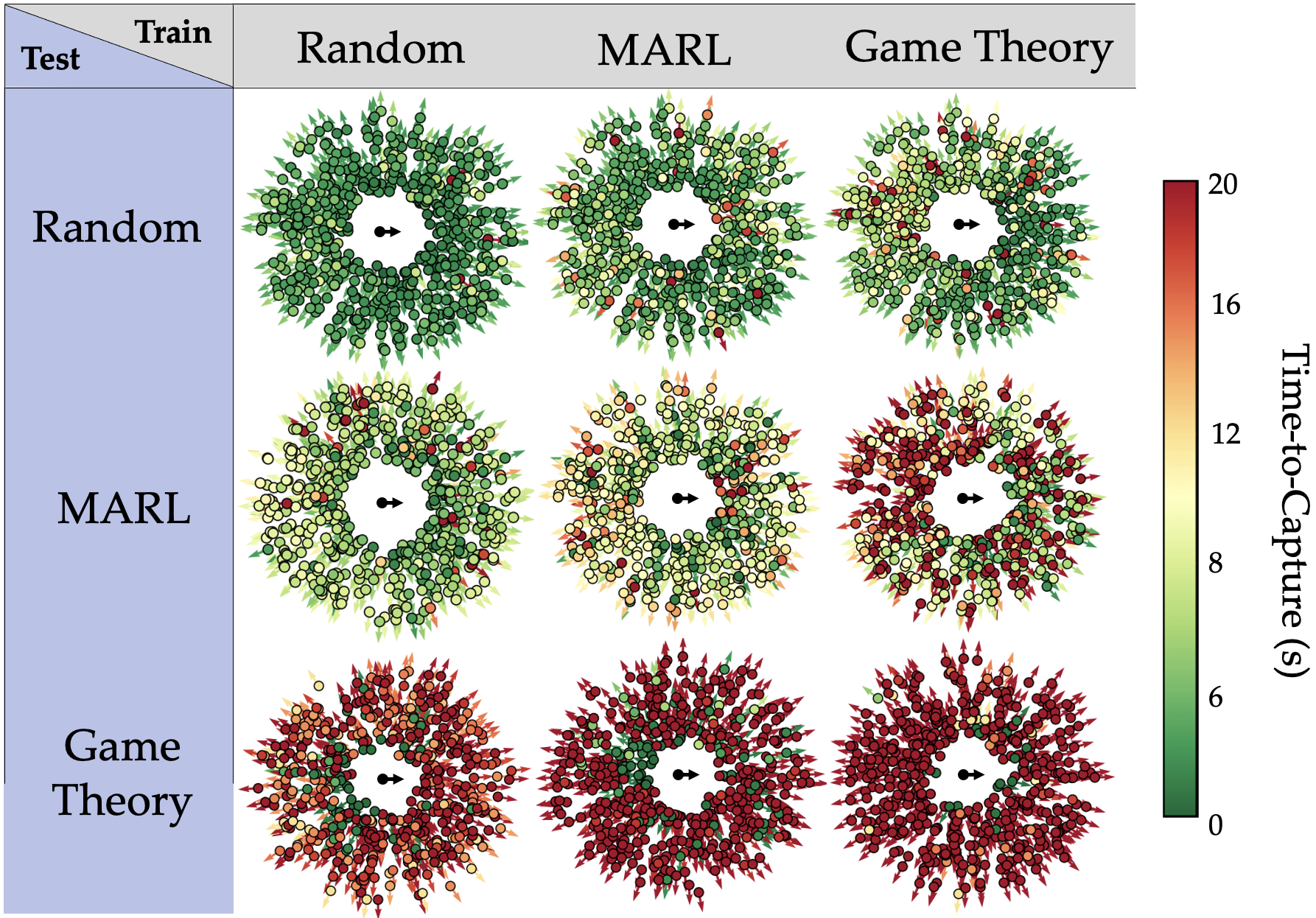}
    \caption{Each \robot policy is trained on a unique distribution of \agent behavior---random, MARL, or zero-sum game-theoretic---and deployed to interact with each other possible \agent. 
    The plots reveal how \robot policies trained on \change{more optimal but less diverse behavior ``overfit'' to those types of maneuvers.
    For example, consider the top-right where the \robot policy is trained against a game theory \agent but tested against the random \agent. The random \agent does not explicitly strategize, but } the capture time distribution of the game-theory policy mirrors the distribution when interacting with the game-theoretic \agent (Fig.~\ref{fig:full_vs_partial_obs}).}
    \label{fig:train_test_performance_table}
\end{figure}

\change{\subsection{Simulation \& training details}
\label{app:sim_and_training_details}
}

\change{\para{Quadruped High \& Low-level Control} The quadruped uses a pre-computed low-level controller for walking, and is just learning a high-level strategic policy. In this way, our method is agnostic to the specific low-level controller. We use an off-the-shelf RL-based approach to train a walking policy \cite{rudin2022learning}. We use this policy during all simulation experiments and during training the high-level strategic policy. 
The high-level policy we present in this work controls linear and angular velocity, $\uR = [v_x^\R, v_\theta^R] \in \mathcal{U}^\R$. The control inputs are bounded: $v_x^\R \in [0, 3] \frac{m}{s}$ and $v_\theta^\R \in [-2, 2] \frac{rad}{s}$. Note that the \robot has a $\sim 17\%$ linear speed advantage over the \agent (detailed in Appendix~\ref{app:evader_design}). }

\change{\para{Optimization Details} The episode length for interactions in the Isaac Gym simulator is always $T = 20 s$. 
The high-level quadruped policy runs at $5 ~Hz$, and simulates a new control applied every $\delta t  = 0.2 s$. The low-level quadruped policy runs at $50 ~Hz$. 
We use the Adam optimizer for both PPO and for DAGGER. We use Exponential Linear Unit (ELU) activations for the policy network and for the encoders. The future state trajectory of the \agent is always $N = 4 s = 8~ \mathrm{steps}$. 
We train the teacher policies with $3,000$ environments, and the student policy with $8,000$ environments. We always train the high level policy on flat terrain in an infinite plane. 
Otherwise, we use all the default parameters from \cite{rudin2022learning} but remove the cross-entropy loss during PPO. }

\change{\subsection{\Agent simulation \& policy design}
\label{app:evader_design} }

\change{We simulate the \agent as a unicycle dynamical system, controlling linear and angular velocity, $\uA = [v_x^\A, v_\theta^\A] \in \mathcal{U}^\A$. The control inputs are bounded: $v_x^\A \in [0, 2.5] \frac{m}{s}$ and $v_\theta^\A \in [-2, 2] \frac{rad}{s}$. The discrete-time dynamics model we use is:
\begin{equation}
    \begin{bmatrix}
        p_x^\A \\
        p_y^\A \\
        p_z^\A \\
        \theta^\A \\
    \end{bmatrix}^{t+\delta t } =
    \begin{bmatrix}
        p_x^\A \\
        p_y^\A \\
        p_z^\A \\
        \theta^\A \\
    \end{bmatrix}^{t} + 
    \delta t \begin{bmatrix}
        v_x^\A \cos(\theta^\A) \\
        v_x^\A \cos(\theta^\A) \\
        0 \\
        v_\theta^\A \\
    \end{bmatrix}.
\end{equation} }

\change{The \agent's initial condition is randomly initialized at the start of each episode and after each environment reset. Specifically, the initial $xy$-position of the \agent is offset from the \robot initial $xy$-position via
\begin{equation*}
    \begin{bmatrix}
        p_x^\A \\
        p_y^\A \\
        p_z^\A \\
        \theta^\A
    \end{bmatrix}^0 = 
    \begin{bmatrix}
        p_x^\R \\
        p_y^\R \\
        0 \\
        0 
    \end{bmatrix}^0 + 
    \begin{bmatrix}
        r^0 \cos(\psi^0) \\
        r^0 \sin(\psi^0) \\
        0.001 \\
        \tan^{-1}(p_y^{\A, 0} / p_x^{\A, 0})
    \end{bmatrix},
\end{equation*}
where $r^0 \sim \mathrm{Unif}[2,6]$ is drawn uniformly at random between 2 and 6 meters, and $\psi^0 \sim \mathrm{Unif}[-\pi, \pi]$ so that the \agent spawns at random angles relative to the \robot. The $z$-position is always fixed at $0.001$ and the yaw angle $\theta^\A$ is chosen so the \agent is facing away from the \robot at the start of the episode. }

\change{\subsubsection{Dubins' policy}
\label{sec:dubins}
This \agent always moves in Dubins' paths \cite{dubins1957curves} with a fixed time duration for turning or going straight determined randomly upon the start of the episode. Let $v^\A_x \in [\underline{v}_x, \bar{v}_x]$ be the min and max linear velocity commands and $v^\A_\theta \in [\underline{v}_\theta, \bar{v}_\theta]$ be the min and max angular velocity commands. Let $\tau_{fwd}$ be the number of seconds for which the \agent goes forward, and $\tau_{turn}$ be the number of seconds to turn. At the start of the episode, each parameter is sampled from the corresponding uniform distribution: 
$$\tau_{fwd} \sim \mathrm{Unif}[2, 4], \quad \tau_{turn} \sim \mathrm{Unif}[0.6, 1.4].$$ 
The Dubins' policy switches between two maneuvers depending on if the current time $t$ matches the time interval $\tau_{fwd}$ or $\tau_{turn}$: 
\begin{itemize}
    \item $\tau_{fwd}$: \agent goes forward at max linear velocity, applying $\uA = [\bar{v}_x, 0]$
    \item $\tau_{turn}$: \agent turns while moving at max linear velocity. With $70\%$ probability the \agent turns the \textit{opposite} direction as the last turn, and with $30\%$ turning the \textit{same} direction as last time): $\uA = [\bar{v}_x, \bar{v}_\theta]$ or $\uA = [\bar{v}_x, \underline{v}_\theta]$
\end{itemize}
At the start of the episode, the \agent uniformly at random chooses to turn left or right. }

\change{\subsubsection{Random policy}
This evader, $\pi^\A_{\mathrm{rand}}(t)$, randomly samples a motion primitive to apply for a fixed number of seconds, before re-sampling another motion primitive. 
The set of motion primitives $\mathcal{MP}$ are the Cartesian product of regularly discretized linear and angular velocities:
\begin{equation*}
    V_x = [\underline{v}_x, 0.5\underline{v}_x, 0, 0.5\bar{v}_x, \bar{v}_x], \quad V_\theta = [\underline{v}_\theta, 0.5\underline{v}_\theta, 0, 0.5\bar{v}_\theta, \bar{v}_\theta], \quad \mathcal{MP} := V_x \times V_\theta 
\end{equation*}
At the start of each episode, the duration for applying a motion primitive is randomly sampled to be between $1-3$ seconds. When it is time to switch motion primitives, $\pi^\A_{\mathrm{rand}}(t) \equiv (v_x, v_\theta) \sim \mathrm{Unif}[\mathcal{MP}]$. }

\change{\subsubsection{MARL policy}
This evader, $\pi^\A_{\mathrm{marl}}(\xrel_t)$, is trained to evade a pre-trained \robot policy, equivalent to a single iteration of \cite{pinto2017robust}. 
The \agent is rewarded for maximizing the distance between the two agents at each timestep, and obtains a termination penalty upon capture:
\begin{equation}
    \textit{Evasion: } r_t = 2 \cdot ||\xA_t - \xR_t||_2^2, \quad \textit{Capture: } r_T = -80 \cdot \mathbbm{1}\{||\xA_T - \xR_T||_2^2 \leq 0.8\}.
\end{equation}
The \agent trains against a pre-trained, fully-observable \robot policy: $\pi^*(\xrel, z_t)$. This \robot was originally trained against a random \agent, $\pi^\A_{\mathrm{rand}}(t)$. 
Since the \robot is pre-trained, its policy is already highly capable of capturing the \agent making it difficult for the MARL agent to learn. 
To tackle this, we use a curriculum set at where at each fixed iteration, the \robot speed in increased by $20\%$. The curriculum is set to $[1800, 2000, 2400, 2800, 3100, 3800]$ iterations. 
\begin{figure}[h!]
    \centering
    \includegraphics[width=0.8\textwidth]{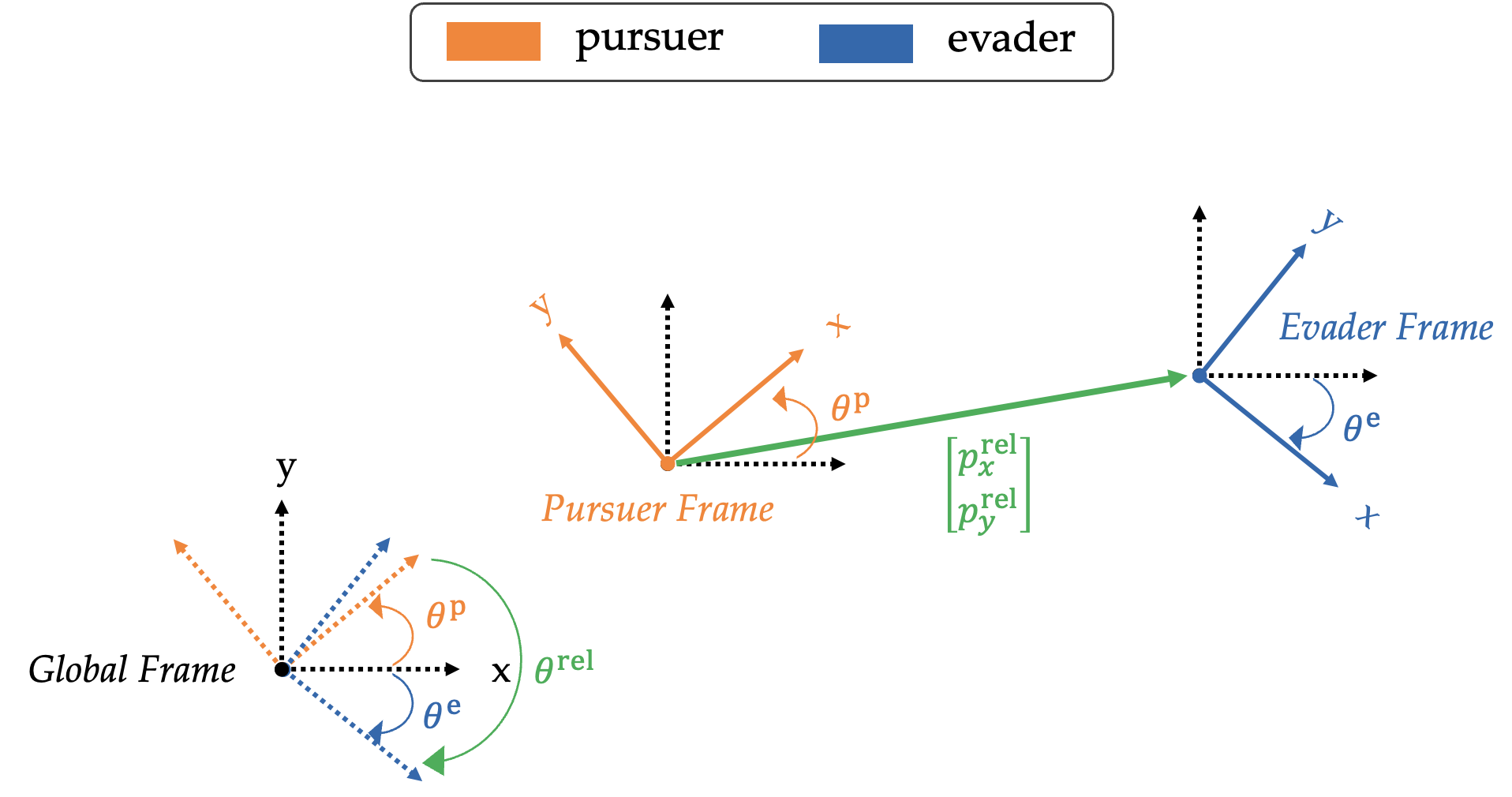}
    \caption{\change{Global and relative coordinate system with respect to the \robot's coordinate frame.}}
    \label{fig:rel_coord_sys}
\end{figure}
}

\change{\subsection{Zero-sum game policies: \robot and \agent}
\label{app:zero_sum_game_solver}
For one of our candidate fully-observable supervisor policies, we use an off-the-shelf zero sum differential game solver \cite{mitchell2005time, mitchell2008flexible}. 
It computes the optimal value of the two-player game, $V(\xrel, t)$ at each time $t \in [0,T]$ via solving the Hamilton-Jacobi Isaacs Partial Differential Equation, whose solution yields optimal game-theoretic policies for both the \robot and the \agent. 
Each agents' state is their global position and their heading angle: $x^i := [p_x^i, p_y^i, \ang^i]^\top, \quad i \in \{\R, \A\}$ (see Fig.~\ref{fig:rel_coord_sys}). 
Each agent controls their linear and angular yaw velocity: $u^i := [v_x^i, v_\ang^i] \in \mathcal{U}^i$ and the dynamics evolve via a unicycle model:
\begin{equation}
\begin{bmatrix}
    \dot{p}_x^i \\ 
    \dot{p}_y^i \\
    \dot{\ang}^i 
\end{bmatrix}
=
\begin{bmatrix}
    v_x^i \cos(\ang^i) \\ 
    v_x^i \sin(\ang^i) \\ 
    v_\ang^i 
\end{bmatrix}.
\end{equation}
}

\change{Each agent's policy is defined in their respective relative coordinate system (see Fig.~\ref{fig:rel_coord_sys} for relative coordinate system in pursuer's body frame). From the perspective of the \robot, the relative state $\xrel := [\pxrel, \pyrel, \threl]^\top$ consists of the relative $xy$-position and heading:
\begin{equation}
    \threl := \ang^\A - \ang^\R, \quad 
    \begin{bmatrix}
        \pxrel \\  
        \pyrel \\
    \end{bmatrix} := 
    \begin{bmatrix}
        \cos(\ang^\R) & \sin(\ang^\R) \\ 
        -\sin(\ang^\R) & \cos(\ang^\R) 
    \end{bmatrix} \begin{bmatrix}
        p_x^\A - p_x^\R \\ 
        p_y^\A - p_y^\R
    \end{bmatrix}
\end{equation}
The continuous-time relative dynamics of the \agent in the \robot's body frame is defined by:
\begin{equation}
    \begin{bmatrix}
    \dpxrel \\
    \dpyrel \\
    \dthrel \\
    \end{bmatrix} = 
    \begin{bmatrix}
        -v_x^\R + v_x^\A \cos(\threl) + v_{\ang}^\R \pyrel \\
        v_x^\A \sin(\threl) - v_\ang^\R \pxrel \\
        v_{\ang}^\A - v_\ang^\R \\
    \end{bmatrix} := f(\xrel, \uR, \uA).
\end{equation}
Note that when solving the game, the dynamics models are assumed to be deterministic. 
}

\change{The game is solved over the same time horizon as the episode length in the Isaac Gym simulator: $T = 20 s$. The \robot's optimal policy is defined as:
\begin{equation}
    \pi^{\R}_{\mathrm{game}}(\xrel, t) := \arg \min_{\uR \in \mathcal{U}^\R} \max_{\uA \in \mathcal{U}^\A} \nabla V(\xrel, t)^\top f(\xrel, \uR, \uA),
    \label{eq:pursuer_opt_game_policy}
\end{equation}
The \agent's optimal policy is defined similarly with two changes: 1) the relative state $\xrel$ is of the \robot's state in the \agent's body frame, and 2) the $\min$ and the $\max$ are swapped in Eq.~\ref{eq:pursuer_opt_game_policy}.  
\begin{equation}
    \pi^{\A}_{\mathrm{game}}(\xrel, t) := \arg  \max_{\uA \in \mathcal{U}^\A} \min_{\uR \in \mathcal{U}^\R} \nabla V(\xrel, t)^\top f(\xrel, \uR, \uA).
\end{equation}
Note that because the joint dynamics are affine in the \robot and \agent controls, the value of the game from the \robot and \agent's perspective is identical (also known as \textit{Isaacs' condition} \cite{isaacs1999differential}). 
}

\change{\para{Supervising the partially-observable policy} Recall that our approach uses supervision from the fully-observable policy. 
The partially-observable policy's intent estimate, $\zhat_t$, is supervised with the teacher's intent encoding $z_t$ and also the \robot actions, $\uR_t$, are supervised with the teacher policy actions, $u^{\R,*}_t$. However, the game-theoretic policy does not generate an explicit latent state $z_t$ the way that the MARL and Random policy architectures do. Thus, we only supervise partially observable policy's with the game theory policy $\pi^{\R, *}(\xrel_t, t)$ at each timestep and we update the latent intent encoder $\mathcal{E}$ and the policy $\pi^\R$ weight's. In the continual learning of Sec.~\ref{subsec:adapting_to_ood} we follow this same paradigm: finetune the weights of the encoder and the policy from just action supervision. Otherwise the architecture of the partially observable student policy is identical (LSTM for encoding with 3-layer MLP for action network). }

\change{\subsection{Kalman Filter}
\label{app:kalman_filter}}

\change{We use a linear state transition model, $$\xrel_{t+\delta t} = A\xrel_t + B\uR_t + w$$ which ignores the role of the \agent (i.e., $\uA_t \equiv 0$) during the the prediction step, and a linear observation model, $$y_t = H\xrel_t + v.$$
Note that while the Kalman filter ignores the motion of the evader, the student policy does not. Indeed, the student policy uses a history of relative states to predict the future evader’s motion (Fig.~\ref{fig:approach}). We intentionally ignore the evader’s motion in the Kalman filter prediction step to waive the requirement for a velocity estimator (which can be cumbersome to obtain and noisy in the real world). While this makes filtering imperfect, we empirically found that the learned policy can cope with the resulting inaccuracies. }

\change{The state transition model assumes the relative state evolves via a single integrator model, assuming $\uR := [v_x, v_y, v_\theta]$, which directly influence the time derivatives of the relative $x, y$ and $\theta$ states. The quadruped we used in the experiments cannot control $v_y$ directly, so at deployment time this entry of the control input is set to zero. 
The process covariance $w \sim \mathcal{N}(0,Q)$, the measurement covariance $v \sim \mathcal{N}(0,R)$, initial error covariance matrix $P_0$, observation matrix $H$, and state transition matricies are: 
\begin{align}
    Q &= 0.01 \cdot I_{3\times3}, \quad R = \mathrm{diag}(0.2, 0.2, 0.1), \quad P_0 = I_{3\times3}, \quad H = I_{3\times3}, \\ 
    A &= I_{3\times3}, \quad B = \mathrm{diag}(-\delta t, -\delta t, \delta t), \quad \delta t = 0.2
\end{align}
}

\change{\subsection{Real-world experiments}}

\change{We use an Unitree A1 robot and a Unitree Go1 robot for experiments. The policy controls the linear speed of the robot on the x-axis in the body frame, i.e., in the forward direction ($v_x$), and the angular yaw velocity ($v_\theta$). Such high-level commands are executed by an off-the-shelf low-level policy~\cite{loquercio2022learn}.
We use a Zed 2 camera for object detection in 3D. The object detection pipeline and the Kalman filter run onboard in a Jetson Tx2, while the low-level controller runs on an Intel UP Board. Communication happens through ROS using the implementation of~\cite{loquercio2022learn}.}

\change{Our policy only controls the pursuer Unitree A1 robot, while the evader is either a human or a Go1 robot teleoperated by a human. In both cases, the evader policy is neither pre-scripted nor explicitly strategic. The evader tries to avoid the pursuer with short-horizon evasion maneuvers, but it is not strategic in a long horizon to keep the interaction spatially constrained.
We switch between tracking a person and another robot by changing the tracking class from PERSON to ANIMAL in the ZED 2 camera configs. 
We assume that only a single evader is present in the scene; therefore, we do not keep track of object identity.
We empirically find that the ANIMAL class has more mis-detections than the PERSON class since the visual features of the quadruped differ from a real dog. Still, the detection performance was sufficient for continuous interaction.
}

\change{\subsection{Section~\ref{subsec:ablation_latent_state} details: Reactive, Lookback, Lookahead policies}}

\begin{figure}
    \centering
    \includegraphics[width=\textwidth]{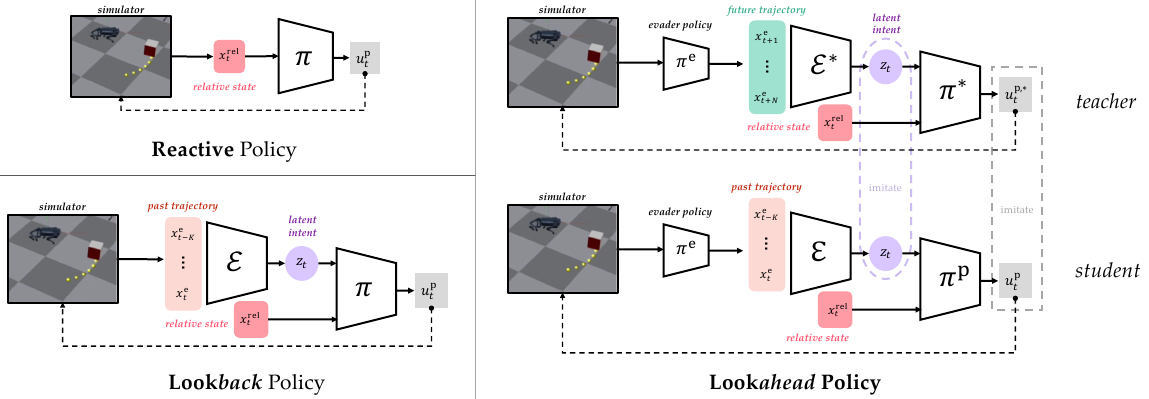}
    \caption{Policy architectures for reactive (up left), lookback (low left) and lookahead policy (right).}
    \label{fig:lb_vs_la_arch}
\end{figure}

\begin{figure}[h!]
    \centering
    \includegraphics[width=\textwidth]{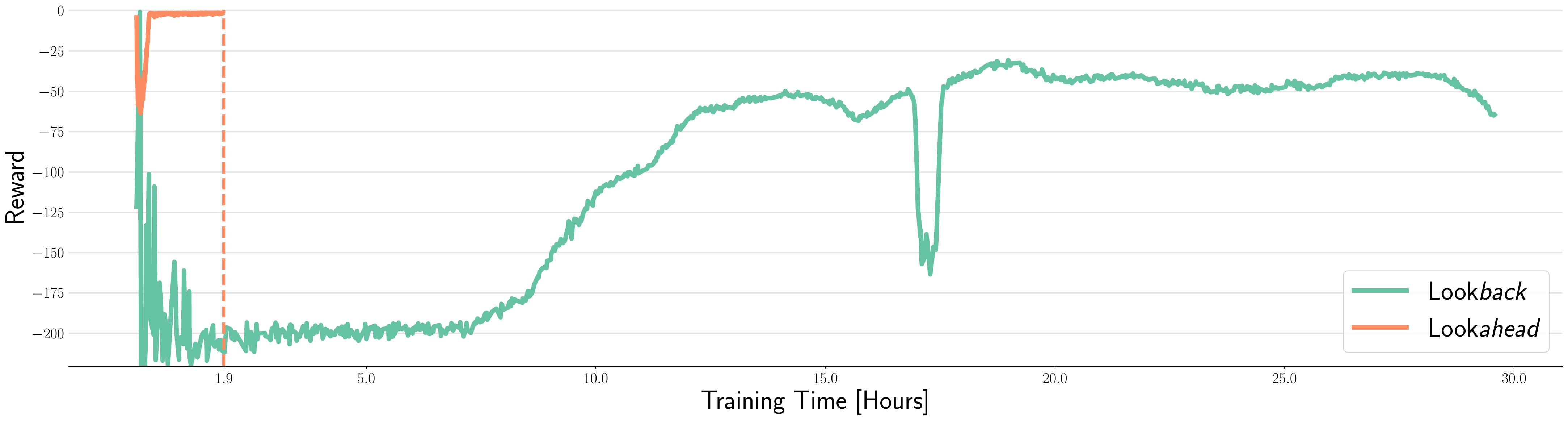}
    \caption{The lookahead policy (orange) converges to a higher reward 10 times faster than the lookback policy (green)}
    \label{fig:lb_vs_la_train}
\end{figure}

\change{All policies use the same simulation setup as in Sec.~\ref{app:sim_and_training_details}. The main differences are in the inputs and in the latent intent learning.  Fig.~\ref{fig:lb_vs_la_arch} visualizes the policy architectures for the \textbf{reactive}, \textbf{lookback} and \textbf{lookahead} policies. 
We also compare training time of our \textbf{lookahead} approach--which uses future trajectory information as privileged info to learn the latent intent--to a \textbf{lookback} policy which must estimate the latent intent only from a history of relative states via an LSTM. 
We see that future trajectory supervision enables 10 times faster training compared to the \textbf{lookback} policy (Fig.~\ref{fig:lb_vs_la_train}) as well as higher overall reward. 
}

\change{\subsection{Note on directly learning a strategic pursuer policy with PPO}}

\change{We tried to directly learn a policy with partial observations, $\pi(\hat{x}_t, \Sigma_t)$, using PPO. However, our policies never got a performance better than random. We believe the main reason to be the difficulty of the optimization problem. Indeed, such a policy needs to learn at the same time: (1) strategy, i.e., learning to predict where the evader will be, (2) control, i.e., learning to move towards a desired location on the x-y plane, and (3) estimation, i.e., learning to reason over the noisy states from the Kalman filter to get an estimate of the real position of the evader. We empirically found that PPO fails to do all these tasks simultaneously. However, we found that PPO could easily solve (1) and (2) if provided with a short-horizon future trajectory of the evader: this corresponds to our fully-observable training.  To reduce sample complexity, we learn (3) using distillation learning, even though it could as well be trained with a further stage of reinforcement learning. An interesting venue for future work is to see how to learn such policy directly.}

\end{document}